\documentclass[10pt,twocolumn,letterpaper]{article}
\usepackage[accsupp]{axessibility}  %

\usepackage{iccv}      %
\usepackage{tabularx}
\usepackage{longtable}
\usepackage{multirow} 
\usepackage{float}
\usepackage{listings}
\usepackage[utf8]{inputenc}
\usepackage[T1]{fontenc}

\newcommand{\topic}[1]{{\vspace{5pt}\noindent\textbf{#1}}}

\definecolor{iccvblue}{rgb}{0.21,0.49,0.74}
\usepackage[pagebackref,breaklinks,colorlinks,allcolors=iccvblue]{hyperref}
\usepackage{courier}

\def\themodel{{{DIMO}}\xspace} 

\def\paperID{5} %
\def\confName{ICCV}
\def\confYear{2025}

\title{\themodel: Diverse 3D Motion Generation for Arbitrary Objects}

\author{Linzhan Mou$^{1}$ \qquad Jiahui Lei$^{1}$$^\dagger$ \qquad Chen Wang$^{1}$ \qquad Lingjie Liu$^{1}$ \qquad Kostas Daniilidis$^{1,2}$ \vspace{0.1em} \\ 
$^{1}$University of Pennsylvania \qquad $^{2}$Archimedes, Athena RC  \vspace{0.1em}\\
}

\begin{document}

\twocolumn[\maketitle
\centering
\vspace{-8mm}
\includegraphics[width=1.0\linewidth]{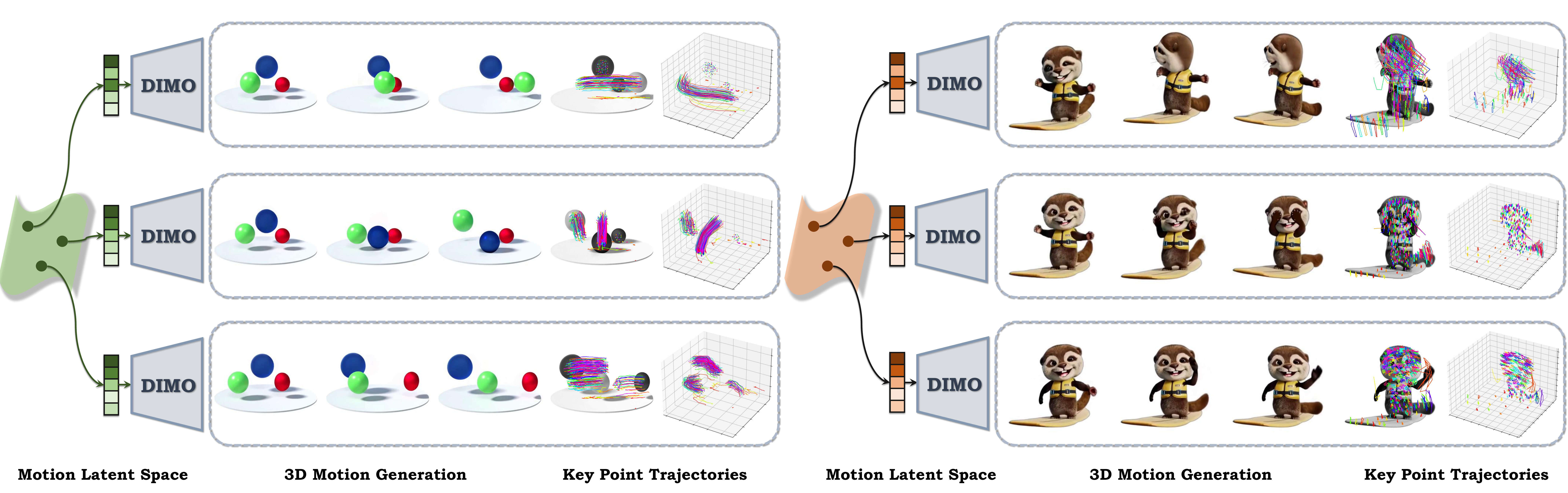}
\captionof{figure}{During inference, \textbf{\themodel} can \underline{\textit{\textbf{instantly}}} generate \underline{\textit{\textbf{diverse}}} 3D motions and high-fidelity 4D contents in \underline{\textbf{\textit{a single forward pass}}} from \underline{\textbf{\textit{a single generative model}}}, by sampling from a continuous motion latent space.}
\vspace{1mm}
\label{fig:teaser}
\bigbreak]
\begin{abstract}
\label{sec:abstract}
We present \textbf{\themodel}, a generative approach capable of generating diverse 3D motions for arbitrary objects from a single image. The core idea of our work is to leverage the rich priors in well-trained video models to extract the common motion patterns and then embed them into a shared low-dimensional latent space. Specifically, we first generate multiple videos of the same object with diverse motions. We then embed each motion into a latent vector and train a shared motion decoder to learn the distribution of motions represented by a structured and compact motion representation, i.e., neural key point trajectories. The canonical 3D Gaussians are then driven by these key points and fused to model the geometry and appearance. During inference time with learned latent space, we can instantly sample diverse 3D motions in a single-forward pass and support several interesting applications including 3D motion interpolation and language-guided motion generation. Our project page is available at \href{https://linzhanm.github.io/dimo/}{https://linzhanm.github.io/dimo}.

\end{abstract}

\vspace{-6mm}
   
\renewcommand{\thefootnote}{\fnsymbol{footnote}}
\footnotetext[0]{$^\dagger$ Corresponding Author \{leijh@cis.upenn.edu\}}
\renewcommand{\thefootnote}{\arabic{footnote}}
\section{Introduction}
\label{sec:introduction}

4D generation should be diverse in terms of motion. Our community desires a generative model capable of producing these dynamic 3D objects, and once generated, with not just a single motion per object but a space of many possible motions. This paper takes the first step toward addressing this challenging problem.

Previously, generating dynamic objects with such rich motion spaces has only been possible for category-specific objects, such as humans and animals, using template models where motion priors are obtained through time-consuming and labor-intensive motion capture. A successful example is the SMPL~\cite{smpl,smplx} model for humans, whose generative capabilities are built upon massive pose sequences captured from real human data. However, this paradigm is not scalable to more general objects in the real world. \textit{How can we build an SMPL-like model that can model and generate diverse 3D motions for any dynamic object?}

On the other hand, previous 4D object generative models~\cite{4dfy,4dgen,dream_in_4d,consistent4d,align_gs,animate124,dreamgaussian4d,mav3d,sc4d,efficient4d,diffusion2,eg4d,4diffusion,dreammesh4d,l4gm,sv4d,sv4d2} can work on general objects, but can only generate one motion per object in each expensive inference pass. Generating diverse 3D motions of the same object requires re-running the diffusion model~\cite{svd,sv3d,sv4d,zero123} and 4D reconstruction~\cite{nerf,3dgs} repeatedly, which will easily lose identity and cost extensive computing. \textit{How can we directly output a 4D object with a diverse motion space that can be sampled instantly during inference? }

Our key insight to address these issues is that recent advanced video models~\cite{svd,cogvideox,sora,wan} already contain rich motion prior knowledge for general objects and can serve the role of the previous expensive motion capture for category-specific objects like humans. With a proper motion representation, we can distill motion priors to specific objects being generated and build an SMPL-like model for any dynamic object, benefiting from jointly learning numerous motion patterns of the same object together.

Building upon this insight, we introduce \themodel, the first generative model of diverse 3D motions for any general objects. \themodel~addresses two key issues: \textit{where} the diverse motions prior knowledge comes from and \textit{how} to model these motions into a unified object-specific motion space. First, to obtain diverse motion knowledge for the target object, we generate numerous videos containing diverse motion patterns from a single-view image with diverse motion text prompts~\cite{llama,gpt4} (Sec.~\ref{subsec:data}). Second, to jointly model diverse motion patterns, we propose to factorize each 3D motion sequence into explicit and compact neural curves represented by sparse key point trajectories (Sec.~\ref{subsubsec:motion}). This allows us to embed motions into a shared low-dimension latent space and jointly learn their diverse distributions within a single generative model (Sec.~\ref{subsubsec:latent}). To further capture the object geometry and appearance, we attach canonical 3D Gaussians~\cite{3dgs} to the dynamic neural key points and fuse them for differentiable 4D optimization using only photometric losses (Sec.~\ref{subsubsec:3dgs}). 

During inference time, we can \textbf{instantly} generate diverse 3D motions and 4D contents in a single forward pass by sampling from the motion latent space. We can also generate new 3D motions by interpolating within this space and reconstruct unseen motions by optimizing latent codes. Additionally, it supports the automatic creation of 4D animations that align with natural language descriptions, making motion generation both intuitive and versatile (Sec.~\ref{subsec:applications}).

In summary, our main contributions include: 
\begin{enumerate} 
    \item We propose the first generative approach of \textit{diverse} 3D motions for any \textit{general} objects from a single-view image, by distilling motion priors from video models. 
    \item We embed each motion pattern into shared latent space and \textit{jointly} learn the diverse motion distributions represented by structured neural key point trajectories.
    \item At inference time, we can \textit{instantly} generate \textit{diverse} 3D motions and 4D contents in \textit{a single forward pass} and support applications like latent space motion interpolation and language-guided motion generation.
    \item State-of-the-art performance on extensive settings and standard 4D generation benchmarks.
\end{enumerate}

\section{Related Works}
\label{sec:related_works}

\begin{figure*}[ht]
    \centering
    \includegraphics[width=1.0\linewidth]{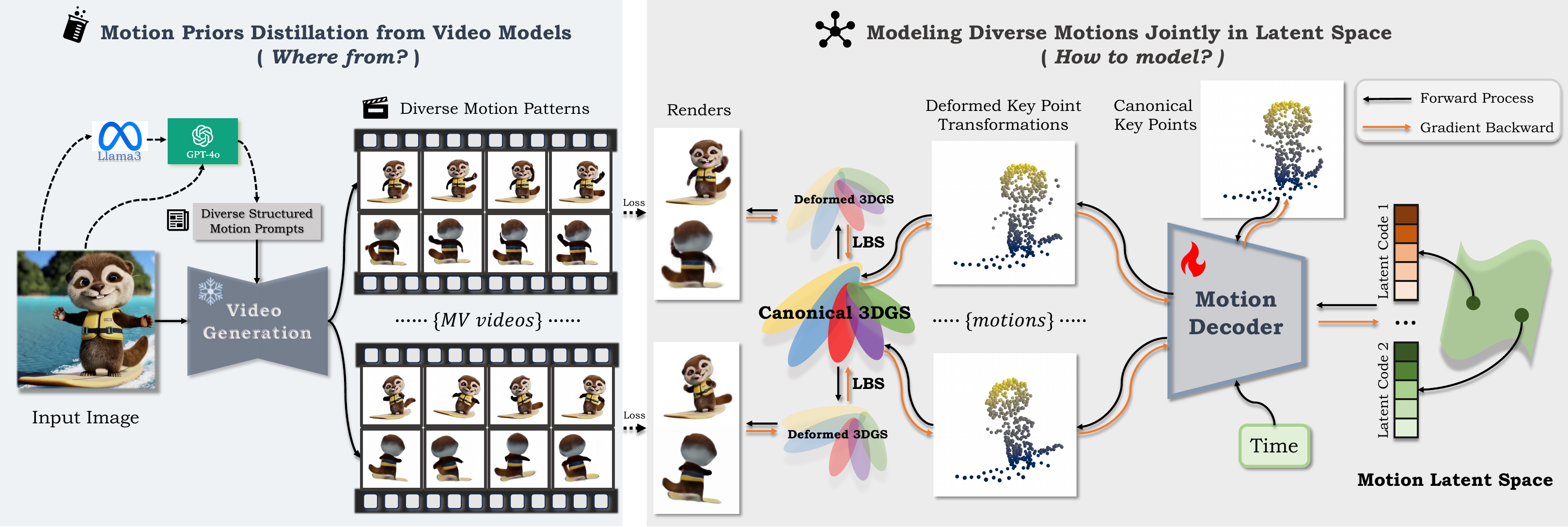}
    \vspace{-7mm}
    \caption{\textbf{Pipeline Overview.} Given a single-view image of any general object, \themodel first \textit{distills rich motion priors} from video models (Sec.~\ref{subsec:data}). We then represent each motion as structured neural key point trajectories (Sec.~\ref{subsubsec:motion}). During training, we embed each motion sequence into a latent code in motion latent space and \textit{jointly model diverse motion patterns} using a shared motion decoder (Sec.~\ref{subsubsec:latent}). The decoded key point transformations are used to drive canonical 3DGS for 4D optimization with only photometric losses (Sec.~\ref{subsubsec:3dgs}).}
    \label{fig:pipeline}
    \vspace{-6mm}
\end{figure*}

\topic{Video Generative Model.} 
Leveraging Internet-scale image and video datasets, 2D video diffusion models (e.g. T2V and I2V)~\cite{video_diffusion,latent_video_diffusion,make_a_video,animatediff,svd,cogvideox,sora,wan,hunyuan} have demonstrated impressive results in generating photo-realistic videos with consistent geometry and plausible motion patterns.
Building on the success of these 2D video generation models, a series of works~\cite{sv3d,vivid123} have adapted the latent video diffusion model as the 3D generator to generate novel views of static objects from different viewpoints by fine-tuning it on 3D data. Other works insert additional modules to enable camera control of video diffusion models~\cite{vd3d,controlling,camco,cvd,ac3d}. Most recent works~\cite{sv4d, 4diffusion,diffusion4d,cat4d} extend this capability into the 4D domain, which leverages the pre-trained video generation model for 4D generation by incorporating an additional view attention layer to align multi-view images.

\topic{Diffusion-Based 4D Generation.} 
Recent diffusion-based 4D generation methods have demonstrated significant advancements in achieving spatio-temporal consistency and motion fidelity. Existing optimization-based approaches~\cite{4dfy,4dgen,dream_in_4d,consistent4d,align_gs,animate124,dreamgaussian4d,mav3d,sc4d,dreammesh4d,dreamscene4d} leverage pre-trained text-to-image~\cite{ldm}, text-to-video~\cite{make_a_video}, and image-to-3D~\cite{zero123} diffusion models to distill a unified 4D representation (deformable neural 3D representation)~\cite{nerf,3dgs} via score distillation~\cite{dreamfusion, prolificdreamer}. In contrast, photogrammetry-based methods~\cite{efficient4d,diffusion2,eg4d,sv4d,sv4d2} directly generate multi-view videos of a 4D object and use them as supervision for subsequent 4D reconstruction. Despite these advancements, current 4D generation approaches still focus on per-motion optimization from a single prompt and fail to generate diverse motion patterns during a single inference stage.

\topic{3D Motion Modeling and Generation.} 
Modeling the motion patterns of dynamic objects is crucial for behavior analysis and content generation. Recent works~\cite{deepphase,temos,actor,unsupervised,dvgan,tevet2023human,zhang2023generating,zhang2024motiondiffuse,wan2023tlcontrol,zhou2024emdm} have explored learning generative models for 3D human motions, leveraging parametric human shape models such as SMPL~\cite{smpl}. While some studies have concentrated on modeling animal motions using keypoint tracking-based methods~\cite{bkind,bkind-3d,mas}, others have learned articulated~\cite{magicpony,farm3d,3dfauna,ponymation} and animatable~\cite{animal_avatars, yang2023reconstructing} 3D animals with generative motion templates. However, 3D motions generated by these models are often restricted to a \textit{specific} category or skeleton structure, relying on category-specific template models or requiring extensive annotated data. 
In contrast, our work aims to model and generate 3D motions for any \textit{general} objects without any pre-defined template model or human-annotated data.

\vspace{-1mm}
\section{Method}
\label{sec:method}
\vspace{-1em}
\topic{Overview.} Given a single-view image of a \textit{general} object, our goal is to model and generate its \textit{diverse} 3D motions. The core idea is to distill the rich motion priors from well-trained video models and embed them into a shared latent space. As shown in~\cref{fig:pipeline}, we first distill diverse motion patterns for the target object from video models (\cref{subsec:data}). Next, we introduce a motion latent space to jointly model the underlying motion distributions (\cref{subsec:model}). To ensure efficient and robust training, we adopt a coarse-to-fine optimization strategy to jointly learn the diverse motion space and complex object geometry (\cref{subsec:optimization}). 

\subsection{Motion Priors Distillation from Video Models}
\label{subsec:data}
A fundamental challenge in diverse 3D motion generation for general objects is the limited motion data from labor-intensive and time-consuming capture~\cite{smpl}. However, current well-trained video models~\cite{sora,cogvideox} have already encoded diverse and plausible motion patterns learned from extensive Internet-scale data. With this insight, we first distill rich motion priors for the target object from these video models.

\topic{Sourcing Diverse Motion Prompts from LLMs.} To generate videos with diverse motions, we first need a collection of motion prompts based on the reference image which embeds a partial knowledge about the object motion we intend to explore. Since this information is incomplete, we rely on prior knowledge within the multi-modal large language model GPT-4o~\cite{gpt4} to generate diverse motion prompts with an auto-prompting technique. To align all motions with the same initial object geometry state, we first employ a fine-tuned Llama3~\cite{llama} to create a structured ``meta'' prompt which contains \textit{a detailed description of appearance, expression, and an initial action state of the object}. Then we feed the reference image and the ``meta'' prompt to GPT-4o and ask for the description of subsequent potential motions.

\topic{Automatic Motion-Rich Video Generation.} 
With motion prompts and reference image based appearance prompts, we use text-conditioned image-to-video model~\cite{cogvideox} to generate motion-rich videos. We use an MLLM~\cite{videoscore} to automatically assess videos and filter out low-quality ones based on a pre-defined score threshold. We also remove those with minimal or excessive motion measured by flow magnitude~\cite{raft}.

\topic{Novel View Priors Distillation.} To enhance the reconstruction generalization, we use multi-view video model~\cite{sv3d,sv4d} to obtain geometric priors by generating object novel views. 

\subsection{Modeling Diverse Motions in Latent Space}
\label{subsec:model}
To jointly model diverse motion patterns within a single generative model, we must have a proper motion representation to effectively model the underlying distributions. Following this, we first model each motion sequence with explicit, compact, and structured key point trajectories. Then we embed motions into the latent space and train a latent-conditioned motion decoder to jointly learn diverse motion distributions. We further attach canonical Gaussians~\cite{3dgs} to capture geometry and fuse them for 4D optimization.

\subsubsection{Key Point Trajectories as Motion Representation}
\label{subsubsec:motion}
Although the geometry and appearance of dynamic objects are often complex and include high-frequency details, the underlying motion that drives these geometries is usually compact (low-rank) and smooth. Inspired by prior works on the Motion Factorization~\cite{lbs,sc-gs,gaussianprediction,dynmf,mosca,shape_of_motion}, we propose to factorize each motion into sparse neural trajectories of key points $\mathcal{P}=\left\{\left(p_k \in \mathbb{R}^3, r_k \in \mathbb{R}^{+}\right)\right\}_{k=1}^{N_k}$ in canonical space, which act as a low-rank motion basis. Specifically, each key point $k$ is parametrized by a canonical position $p_k$ and a global control radius $r_k$, which parameterizes a radial basis function (RBF) that describes its influence weight $w_{jk} \in \mathbb{R}^{+} $ on the nearby point $p_j$:
\begin{equation}
    w_{jk}=\operatorname{Normalize}
    \left(\exp 
        \left(
            -\left\|p_j-p_k\right\|_2^2 / 2 r_k
        \right) 
    \right).
\label{eq:weight}
\end{equation}  

With the low-rank motion basis $\mathcal{P}$, then each motion sequence can be parametrized as neural curves formed by the key points' 6DoF transformations $\mathcal{E}=\left\{\mathcal{E}_k^t\right\}_{t=1}^T$, $k \in\{1, \cdots, N_k\}$, where each pose $\mathcal{E}_{k}^t \in \mathbf{SE}(3)$ at a timestamp $t$ consists of a 3DoF rigid translation $\mathbf{T}_k^t \in \mathbb{R}^3$ and 3DoF rotation quaternion $\mathbf{R}_k^t \in \mathbf{SO}(3)$ of each key point $k$. Formally, each key point based neural curve $\mathbf{c}_k$ is defined as:
\begin{equation}
    \mathbf{c}_k=\left(\left[\mathcal{E}_k^{1}, \mathcal{E}_k^{2}, \mathcal{E}_k^{3}, \ldots, \mathcal{E}_k^{T}\right], r_k\right),
\end{equation}
To maintain the correct topology correlation, we construct the motion graph that connects key points based on their neural trajectories. We define the graph edge $\Omega_k$ as:
\begin{equation}
    \begin{aligned}
    & \Omega_k=\mathrm{KNN}_{j \in\{1, \ldots, N_k\}}\left[ d(p_j^{traj}, p_k^{traj}) / T \right], \\
    & p_k^{traj} = \operatorname{concat}(p_k^1, p_k^2, \cdots, p_k^T)
    \end{aligned}
\end{equation}
where $d(\cdot, \cdot)$ is the Euclidean distance function, $T$ defines the motion sequence length and KNN denotes the K-nearest neighbors under the distance between two trajectories that capture the global topological changes across time.

\subsubsection{Joint Learning of Diverse Motion Distributions} 
\label{subsubsec:latent}
To model the time-varying deformation, previous 4D reconstruction and generation works train a specific deformation network to overfit a single motion, which is time-consuming and not generalizable to diverse motion joint modeling. Most importantly, it doesn't leverage the common motion patterns of the target object. Therefore, to jointly learn diverse motion distributions, we embed a wide variety of motions of the same object into a low-dimensional latent space with latent code $\boldsymbol{z}$ and train a shared deformation network.

Formally, for the motion indexed by $m \in \{1, \cdots, N_m\}$, we employ a latent-conditioned motion decoder $\mathcal{D}_c$ to condition on its latent code $\boldsymbol{z}_m$ and query canonical location $p_k^t$ of each key point $k$ with timestep $t$. The motion decoder $\mathcal{D}_c$ then outputs key point motion-specific 6DoF transformation $(\mathcal{E}_k^t)_m = \left(\mathbf{R}_k^t \mid \mathbf{T}_k^t\right)_m \in \mathbf{SE}(3)$:
\begin{equation}
     (\mathcal{E}_k^t)_m = \mathcal{D}_c \left(\boldsymbol{z}_m, p_k^t, t\right),
\end{equation}
In motion latent space, we assume the prior distribution over each latent vector $\mathrm{p} \left(\boldsymbol{z}_m\right)$ to be a non-zero-mean multi-variate Gaussian as $\boldsymbol{z}_m \sim \mathcal{N}\left(\boldsymbol{\mu}_m, \boldsymbol{\sigma}_m\right)$ with \textit{learnable} parameters $(\boldsymbol{\mu}_m, \boldsymbol{\sigma}_m)$. Then motion-associated vector $\boldsymbol{z}_m$ can be sampled using VAE reparameterization trick~\cite{vae}: 
\begin{equation}
      \boldsymbol{z}_m=\boldsymbol{\mu}_m+\boldsymbol{\sigma}_m \odot \boldsymbol{\epsilon}, \,
     \boldsymbol{\epsilon} \sim \mathcal{N}(0, \boldsymbol{I})
\end{equation}
The learnable distribution parameters $(\boldsymbol{\mu}_m, \boldsymbol{\sigma}_m)$ are learned directly through standard back-propagation. 

\subsubsection{Geometric Modeling with Canonical 3DGS} 
\label{subsubsec:3dgs}
The compact neural trajectories and latent space have effectively modeled the various underlying motions across time. We further employ a set of canonical 3D Gaussians~\cite{3dgs} to capture the object geometry and appearance. Formally, an object is parameterized as Gaussians in canonical space:
\begin{equation}
    \mathcal{G}=\left\{\left(\mu_i, R_i, s_i, o_i, c_i \right)\right\}_{i=1}^{N_g}
\end{equation}
where $\mu_i, R_i, s_i, o_i, c_i$ are the center, rotation, scale, opacity and spherical harmonics (SH) of the Gaussian primitive $i \in \{1, \cdots, N_g\} (N_k \ll N_g)$. To get the geometry at the timestep $t$, canonical Gaussians will be deformed by the nearest canonical key point $p_k$ and its graph edges $\Omega_k$ using widely-used linear blend skinning (LBS)~\cite{lbs} with weighting factor $w \in \mathbb{R}^{+}$ calculated in ~\cref{eq:weight}. 
\begin{equation}
    \begin{aligned}
        \mathcal{G}(t) &= \left\{\left(\mu_i^t, R_i^t, s_i, o_i, c_i \right)\right\}_{i=1}^{N_g} \\
        (\mu_i^t, R_i^t) & = \operatorname{LBS}\left(\left\{w_{ij}, \mathbf{R}_j^t, \mathbf{T}_j^t\right\}_{j \in \Omega_k}\right) (\mu_i, R_i)
    \end{aligned}
\end{equation}
The deformed 3D Gaussians $\mathcal{G}(t)$ are fused and then rendered via a splatting-based differentiable rasterization~\cite{3dgs}.

\subsection{Motion-Oriented Optimization} 
\label{subsec:optimization}
The total learnable parameters include diverse latent codes $\boldsymbol{z}$, canonical key points $\mathcal{P}$, canonical 3D Gaussians $\mathcal{G}$ and a motion decoder $\mathcal{D}_c$. For training efficiency and stability, we adopt a coarse-to-fine motion pre-training schedule.

\begin{figure*}[ht]
    \centering
    \includegraphics[width=1.0\linewidth]{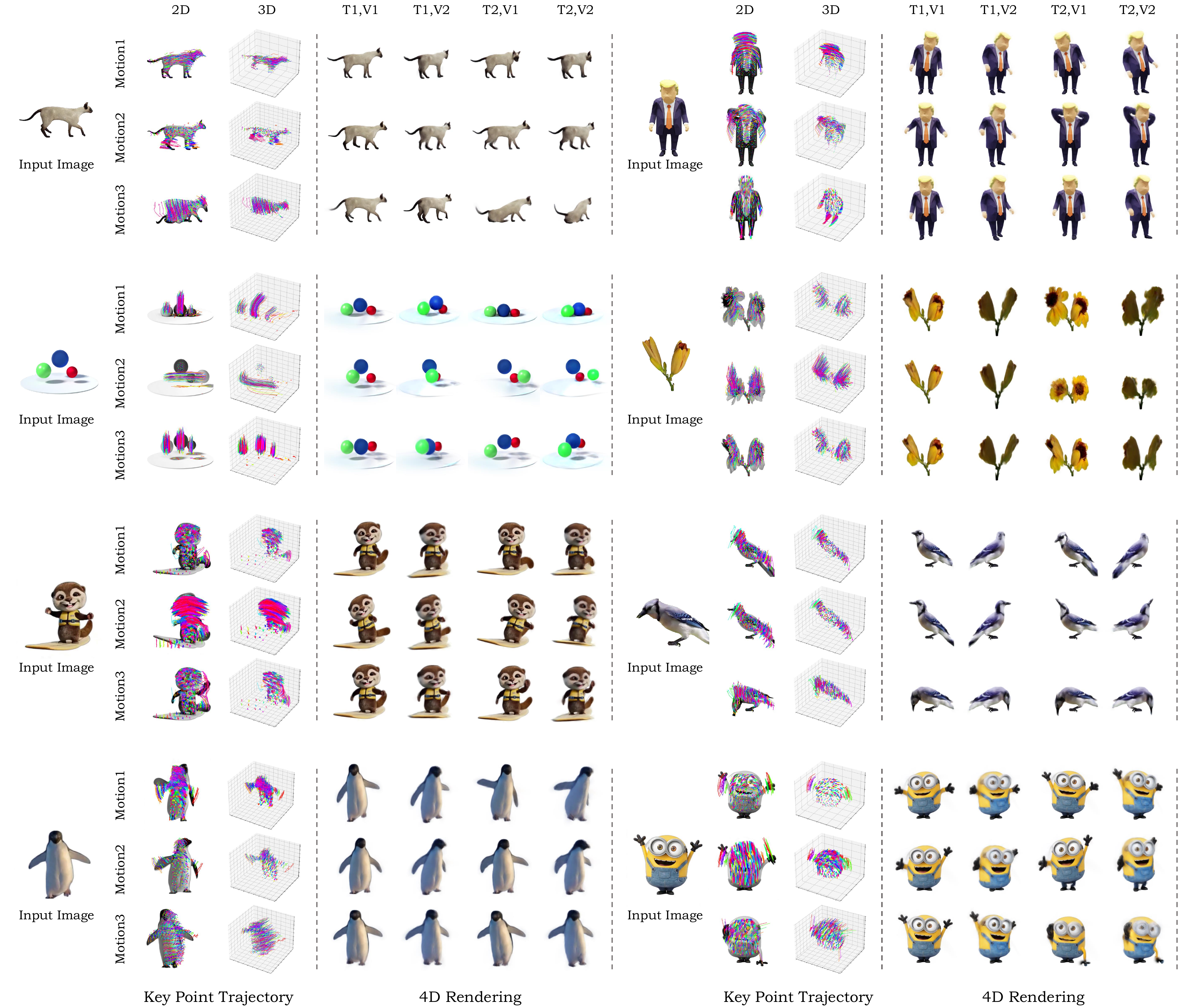}
    \vspace{-5mm}
    \caption{\textbf{Qualitative Results.} During inference, \themodel can \underline{\textbf{\textit{instantly}}} generate \underline{\textbf{\textit{diverse}}} 3D motions and photorealistic 4D contents in \underline{\textbf{\textit{a single forward pass}}} by sampling from latent space. We render three motions for each case under two views at two novel timestamps.}
    \label{fig:results}
    \vspace{-5mm}
\end{figure*}

\topic{Rendering-Based Photometric Optimization.} We randomly sample the latent code $\boldsymbol{z}_m$ and infer the deformed $\mathcal{G}(t)$ for target motion $m$ at timestamp $t$. Then, we render frames at training viewpoints and compare them with the video supervision using RGB loss $\mathcal{L}_\mathrm{rgb}$, mask loss $\mathcal{L}_\mathrm{mask}$ and perceptual LPIPS loss $\mathcal{L}_\mathrm{lpips}$. To encourage smooth 3D surfaces, we also involve edge-aware depth smoothness loss $\mathcal{L}_\mathrm{depth}$ and bilateral normal smoothness loss $\mathcal{L}_\mathrm{normal}$~\cite{sv3d}.
\begin{equation}
    \mathcal{L}_\mathrm{photo} = \mathcal{L}_\mathrm{rgb} + \mathcal{L}_\mathrm{mask} + \mathcal{L}_\mathrm{lpips} + \mathcal{L}_\mathrm{depth} + \mathcal{L}_\mathrm{normal}
\end{equation}

\topic{ARAP-Based Geometric Optimization.} To regularize the 3D motion of unconstrained key points, we optimize an \textit{As-Rigid-As-Possible} (ARAP)~\cite{arap} loss $\mathcal{L}_\mathrm{arap}$ to encourage the pairs of nearby key points ($p_j$ within the graph edges $\Omega_k$ of $p_k$) to be \textit{locally} rigid over time. Formally, given two timestamps separated by interval $\Delta t$, we define $\mathcal{L}_\mathrm{arap}$ as:
\begin{equation}
    \mathcal{L}_{\operatorname{arap}}=\sum_{k=1}^{N_k}\sum_{t=1}^{T}\sum_{j \in \Omega_k} w_{jk} \left\| d\left(p_j^t,p_k^t\right)-d\left(p_j^{t+\Delta t},p_k^{t+\Delta t}\right)\right\|_1
\end{equation}
where $w_{jk}$ is an RBF skinning weight calculated in \cref{eq:weight}.

\topic{Latent Distribution Regularization.} We regularize the latent distribution by minimizing the Kullback–Leibler (KL) divergence $\mathcal{L}_\mathrm{KL}$ between the learned motion distribution $\mathcal{N}(\boldsymbol{\mu}_m, \boldsymbol{\sigma}_m)$ and a standard Gaussian distribution $\mathcal{N}(0, \boldsymbol{I})$:
\begin{equation}
    \label{eq:kl}
    \mathcal{L}_{\mathrm{KL}}=\sum_m^{N_m}-\frac{1}{2}\left(\log \boldsymbol{\sigma}_m-\boldsymbol{\sigma}_m-\boldsymbol{\mu}_m^2+1\right)
\end{equation}

\topic{Coarse-to-Fine Motion Pre-training Schedule.} 
To ensure robust and efficient training, we disentangle the 3D motion and geometry, and adopt a two-stage coarse-to-fine training schedule.
In the \textit{first} stage, we obtain a \textit{coarse} motion basis and latent space by pre-training the latent codes $\boldsymbol{z}$, canonical key points $\mathcal{P}$, and motion decoder $\mathcal{D}_c$. Specifically, we initialize $N_k$ key points as 3D Gaussians within a sphere. During optimization, densification and pruning are performed following 3DGS~\cite{3dgs}. To avoid the local motion minima, every certain iteration we downsample the key points to $N_k$ using FPS~\cite{pointnet++} as an annealing process. After this stage, we obtain a desirable key point distribution in canonical space, which acts as the motion basis for subsequent deformation. More importantly, the model now has learned a reasonable latent space for all motion patterns, on top of which joint optimization of geometry and 3D motion is much more efficient and robust.

In the \textit{second} stage, we further incorporate the canonical 3D Gaussians $\mathcal{G}$ to capture the object geometry and jointly optimize all parameters for \textit{fine-grained} motions. To inherit the canonical shape (distribution) modeled in the first stage, we randomly initialize $n_g$ canonical Gaussians within a small sphere around each canonical key point with radius $r_k$ following~\cite{dreamgaussian,dreamgaussian4d,sc4d}. To improve training stability, we recycle the key point trajectories $\mathcal{E}$ from the first stage to guide the prediction $\hat{\mathcal{E}}$ in the second stage using a Chamfer Distance regularization defined as $\mathcal{L}_\mathrm{chamfer} = \mathrm{CD}\left(\hat{\mathcal{E}}_t, \mathcal{E}_t\right)$. For training efficiency and stability, we gradually increase render resolution from 128×128 to 512×512~\cite{letoccflow}. At each iteration, we randomly sample 4 motions $\times$ 3 views $\times$ 2 frames within a batch, providing multi-motion, multi-view and multi-frame constraints to guide gradient optimization.

\begin{figure*}[ht]
    \centering
    \includegraphics[width=1.0\textwidth]{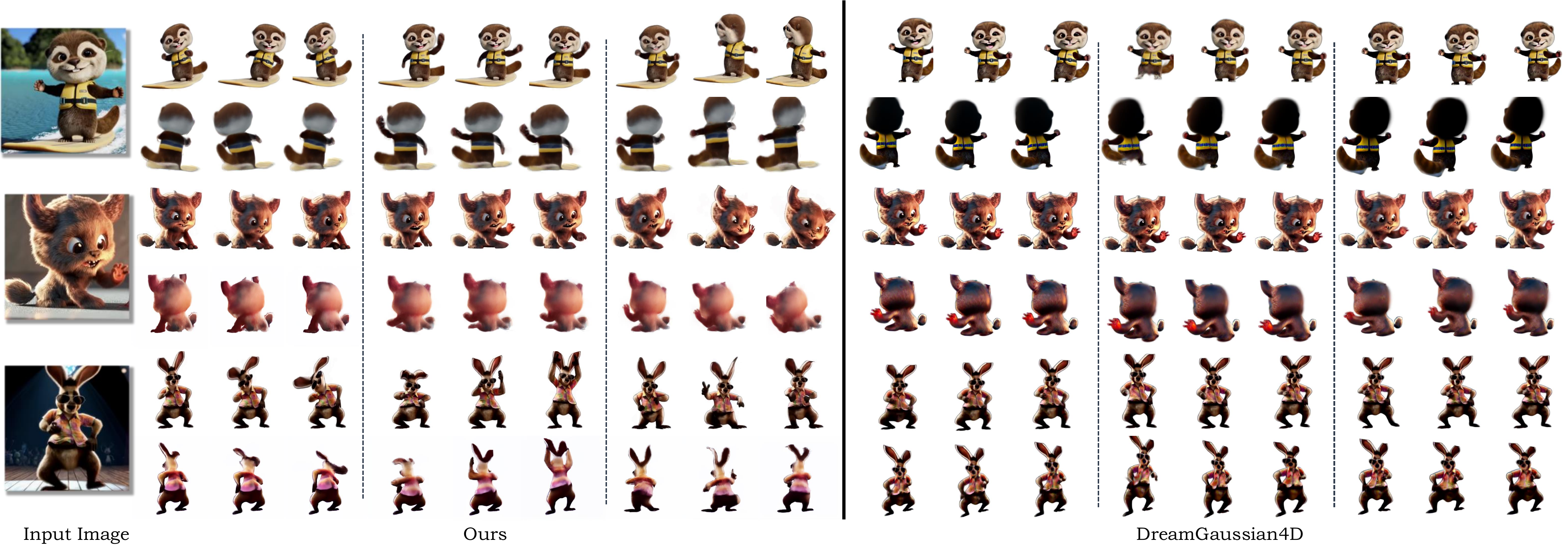}
    \vspace{-7mm}
    \caption{\textbf{Visual Comparison on 3D Motion Generation.} \themodel can generate diverse and high-fidelity 3D motions, whereas the baseline fails to produce noticeable motions (DG4D-generated kangaroo is a thin slice so the back side appears as a mirror image of the front).}
    \label{fig:qualitative}
\end{figure*}
\begin{figure*}[h]
    \centering
    \vspace{-4mm}
    \includegraphics[width=0.9\linewidth]{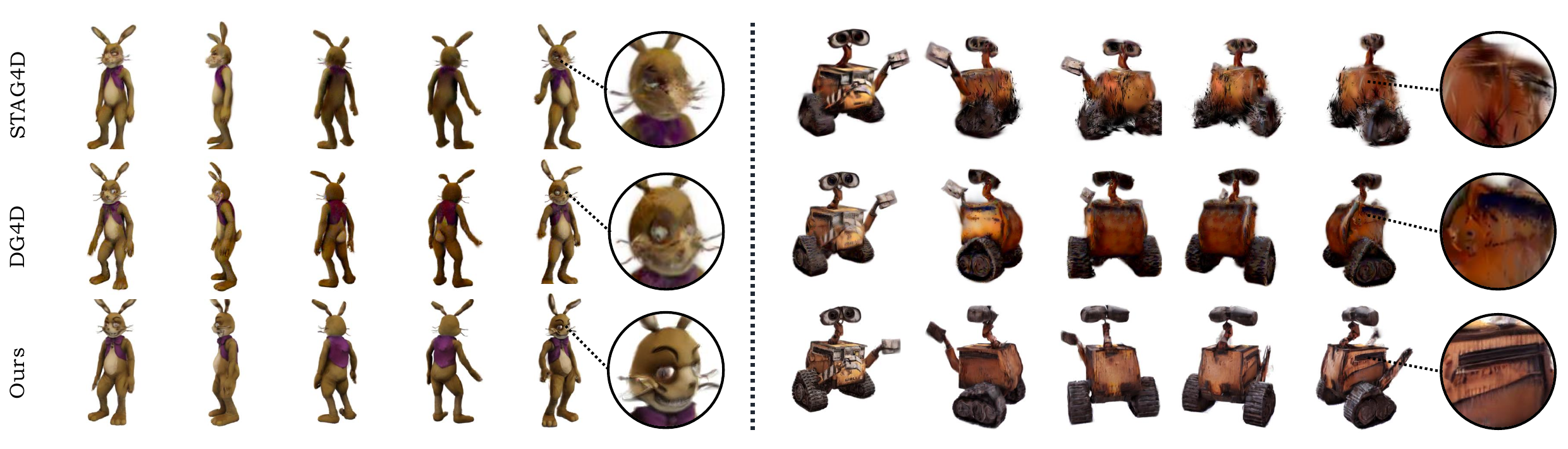}
    \vspace{-4mm}
    \caption{\textbf{Visual Comparison on Image-to-4D.} \themodel can generate high-quality 4D contents for both synthetic and in-the-wild objects.}
    \label{fig:image-to-4d}
    \vspace{-15pt}
\end{figure*}

\begin{figure*}[htbp]
\centering
\begin{minipage}{1.0\linewidth}
    \centering
    \includegraphics[width=0.9\linewidth]{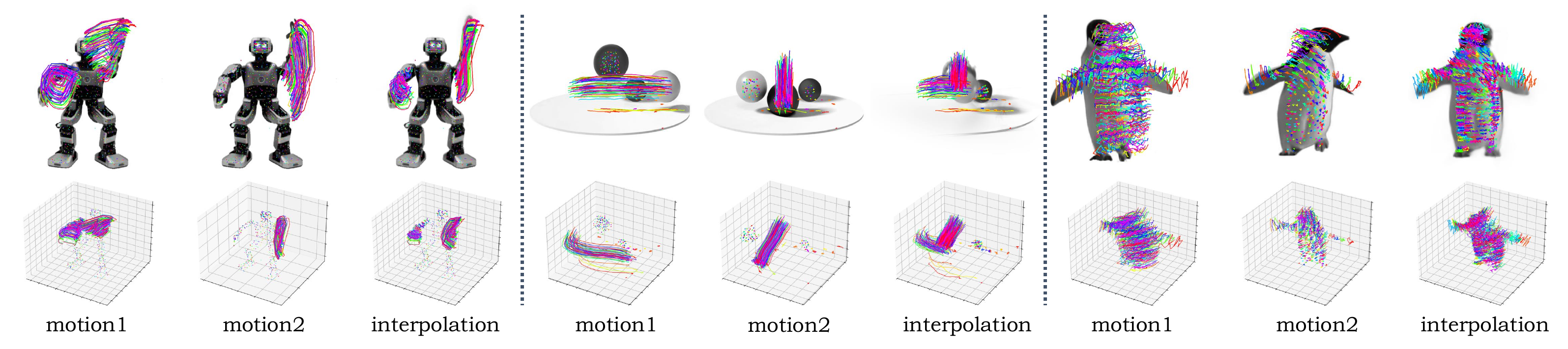}
    \caption{\textbf{3D Motion Interpolation.} We generate new motion by linearly interpolating between two motions sampled from latent space.}
    \label{fig:interpolation}
    \vspace{3mm}
\end{minipage}
\begin{minipage}{1.0\linewidth}
    \centering
     \includegraphics[width=0.9\linewidth]{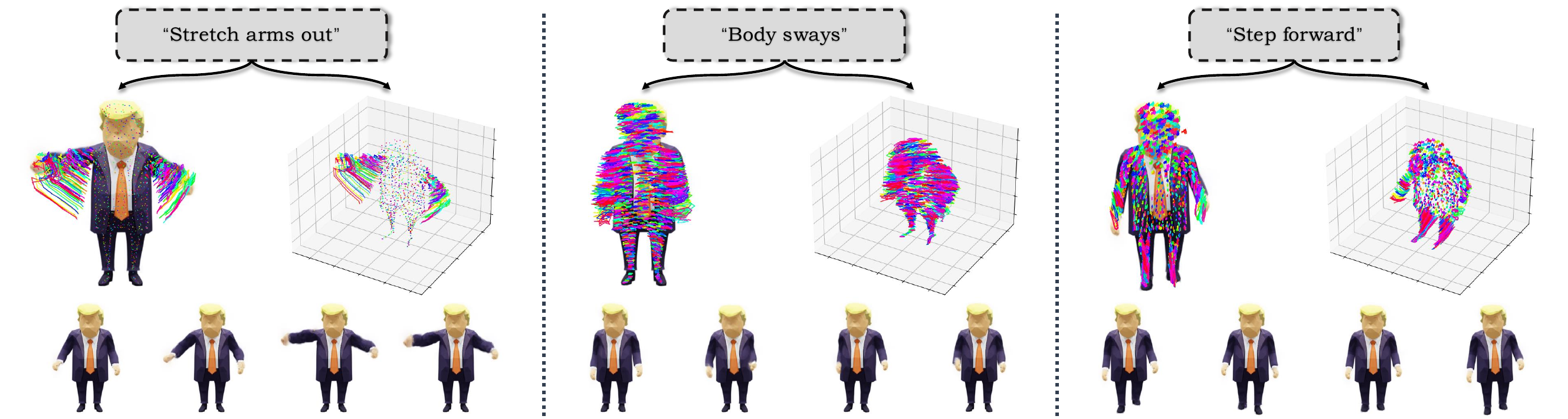}
    \caption{\textbf{Language-Guided Motion Generation.} We project language into latent code and enable feed-forward 3D motion generation.}
    \label{fig:language}
    \vspace{1mm}
\end{minipage}
\vspace{-2mm}
\begin{minipage}{1.0\linewidth}
    \centering
    \includegraphics[width=0.9\linewidth]{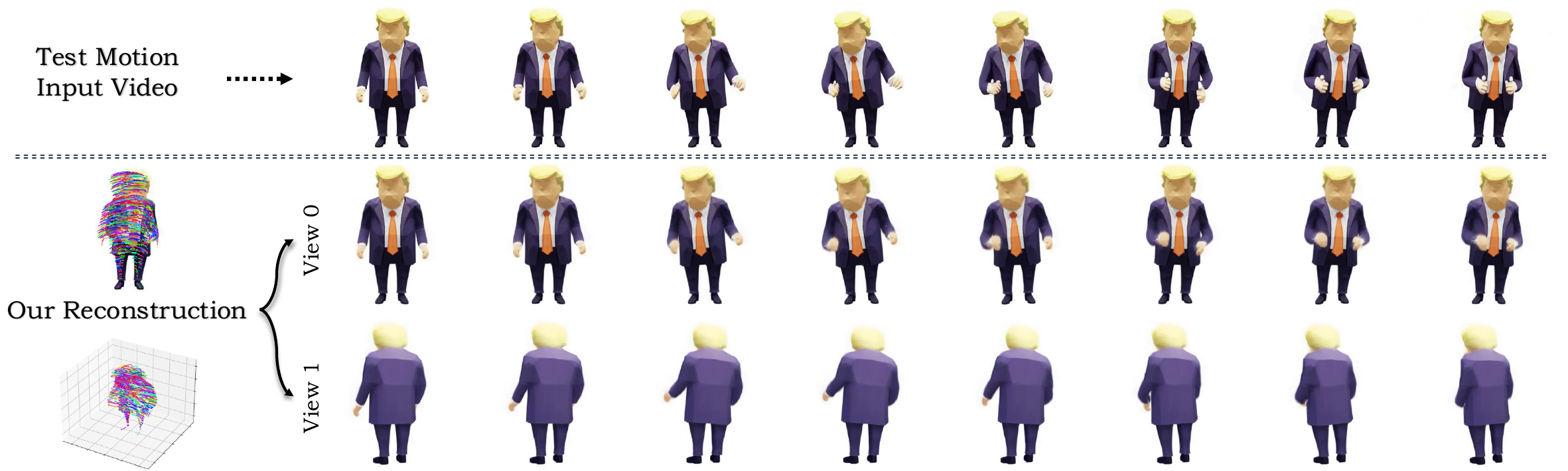}
    \caption{\textbf{Test Motion Reconstruction.} The first row is the input video of the test motion, and the rest are our reconstruction results.}
    \label{fig:test_motion}
    \vspace{-3mm}
\end{minipage}
\end{figure*}

\topic{Language-Guided Motion Generation.} 
\label{subsec:lang}
Since the videos are generated from text prompts, the learned motion latent space should also be compatible with language embeddings. To establish the relationship between natural language and motion space, we first encode the text prompts into text embeddings with pre-trained BERT~\cite{bert} encoder and then train a lightweight MLP to project the text embedding into the motion latent code. 
After this, users could directly provide novel text prompts and our model can generate the corresponding motion and 4D content in a feed-forward manner, as indicated in \cref{fig:language}. Note that to allow people to use simple prompts during inference, we use ChatGPT~\cite{gpt4} to summarize the detailed video caption we use for video generation into a simple motion description such as ``lift the right hand" before feeding them into BERT~\cite{ip2p,instruct4d}.

\vspace{-2mm}
\section{Experiments}
\label{sec:experiments}

We conducted experiments in standard settings and benchmarks to demonstrate the effectiveness of \themodel~in generating diverse and high-fidelity 3D motions and 4D contents. Furthermore, we highlight our applications in motion interpolation, language-guided motion generation, and test motion reconstruction by learning a motion latent space.

\vspace{-1mm}
\subsection{Experimental Setup}
\vspace{-1.9mm}
\label{subsec:setup}
\topic{Dataset and Object Diversity.} We use Objaverse~\cite{objaverse}, Animate124~\cite{animate124}, Consistent4D~\cite{consistent4d}, DAVIS~\cite{davis}, SORA~\cite{sora}-generated and self-collected datasets for experiments. For object diversity, we select a wide range of representative \textit{synthetic} and \textit{real-world} objects including humans, robots, bipods, quadrupeds, birds, plants, fluid and deformable objects with a total of 58 species, each with over 50 motions. 

\vspace{-7pt}
\topic{Implementation Details.} We leverage Llama3~\cite{llama} and GPT4o~\cite{gpt4o} to generate $N_m \ge 50$ diverse motion prompts for each object and adopt CogVideoX5B-I2V~\cite{cogvideox} to generate motion-rich videos. During training, we use 8 views $\times$ 21 frames for each motion sequence and set the virtual camera FOV to $33.9^\circ$ with a fixed radius of $2m$. We use $N_k=512$ key points as the motion basis shared by all motion patterns. For the motion latent space, we employ the Gaussian distribution parameters $(\boldsymbol{\mu}_m, \boldsymbol{\sigma}_m)$ as learnable latent variables for each motion $m$, with a latent dimension of 32. The motion decoder $\mathcal{D}_c$ is implemented as an 8-layer MLP with a skip connection at the 4th layer. The MLP receives motion latent codes, the sinusoidal positional embeddings~\cite{nerf} of time and key points' canonical positions as input, predicting key points' 6DoF transformations. We train a separate generative model for each object. For the 50-motion joint training setting, we pretrain the motions for 2.8k steps in the first stage, which takes roughly 40 minutes on a single 40GB A100 GPU, and for another 8k steps in the second stage for joint optimization of geometry and motion with an additional 3 hours. The rendering speed is 250 FPS at 512$\times$512 resolution with $\sim$80k canonical 3DGS.

\subsection{Comparisons}
\vspace{-1em}
\label{subsec:comparison}
\topic{Comparison on 3D Motion Generation.}
Our method is the first work to generate diverse 3D motions for any general dynamic objects. To evaluate the diversity and quality of our generated 3D motions, we compare our \themodel with DG4D~\cite{dreamgaussian4d} and 4DGen~\cite{4dgen}. We repetitively run their video diffusion models 50 times and perform 4D optimization separately. Then we randomly sample a subset for evaluation. Specifically, we select 7 cases including Human, Cat, Bird, Robot, and SORA~\cite{sora}-generated Kangaroo, Otter, Monster. Each object has five motions. While baselines train each motion sequence separately, our \themodel jointly models all motions in a single generative model. A qualitative comparison of 3D motion generation is shown in \cref{fig:qualitative}. Following~\cite{dreamgaussian4d,animate124,4dgen}, we conduct a user study with 35 participants who are asked to evaluate based on four criteria: motion diversity (MD), image alignment (IA), motion quality (MQ), and 3D appearance (3D App.). The results in \cref{tab:comparison} indicate that our \themodel has a clear advantage over all baselines across all criteria, achieving better 3D motion diversity and visual fidelity.
\begin{table}[H] \scriptsize
    \renewcommand{\arraystretch}{1.2}
    \centering
    \vspace{-3mm}
    \setlength\tabcolsep{8pt}
    \caption{Quantitative comparison on 3D Motion Generation.}
    \vspace{-3mm}
    \label{tab:comparison}
    \begin{tabular}{ p{1.4cm}<{\centering} | p{1.cm}<{\centering} p{1.cm}<{\centering} p{1.cm}<{\centering} p{1.1cm}<{\centering} }
        \toprule
        \textbf{Method} & \textbf{MD}$\uparrow$ & \textbf{IA}$\uparrow$ & \textbf{MQ}$\uparrow$ & \textbf{3D App.}$\uparrow$ \\
        \hline 
        DG4D~\citep{dreamgaussian4d} & 11.4\% & 17.2\% & 8.6\% & 14.3\% \\
        4DGen~\cite{4dgen} & 8.6\% & 20.0\% & 22.8\% & 25.7\% \\
        \textbf{Ours} & \textbf{80.0\%} & \textbf{62.8\%} & \textbf{68.6\%} & \textbf{60.0\%} \\
        \bottomrule
    \end{tabular}
    \vspace{-3mm}
\end{table}

\topic{Comparison on Image-to-4D.} To quantitatively evaluate our generation results, we compare our method with Animate124~\cite{animate124}, 4DGen~\cite{4dgen}, STAG4D~\cite{stag4d}, DG4D~\cite{dreamgaussian4d} and SV4D~\cite{sv4d} on Animate124 and Objaverse dataset. Following~\cite{animate124,4dgen,dreamgaussian4d}, we quantify image alignment using CLIP-I and temporal coherence using CLIP-F. We conduct the user study to measure the motion diversity. The qualitative comparisons are presented in \cref{fig:image-to-4d}. The quantitative comparison results in \cref{tab:image-to-4d} show that our method outperforms the baselines by a significant margin, demonstrating superior performance in visual quality and temporal consistency.
\begin{table}[H]\scriptsize
    \renewcommand{\arraystretch}{1.2}
    \vspace{-3mm}
    \centering
    \caption{Quantitative comparison on Image-to-4D generation.}
    \label{tab:image-to-4d}
    \vspace{-3mm}
    \setlength\tabcolsep{8pt}
    \fontsize{6}{6}\selectfont
    \begin{tabular}{p{1.2cm}<{\centering} | p{0.9cm}<{\centering} p{0.9cm}<{\centering} | p{0.9cm}<{\centering} | p{1.5cm}<{\centering}}
        \toprule
        \textbf{Method} & \textbf{CLIP-I} $\uparrow$ & \textbf{CLIP-F} $\uparrow$ & \textbf{MD} $\uparrow$ & \textbf{Training Time} $\downarrow$ \\
         \midrule
         Animate124~\cite{animate124} & 0.8596 & 0.9756 & 8.1\% & 4.5h \\
         4DGen~\cite{4dgen} & 0.9011 & 0.9854 & 5.4\% & 1h \\
         STAG4D~\cite{stag4d} & 0.9281 & 0.9868 & 8.1\% & 1.5h \\
         DG4D~\cite{dreamgaussian4d} & 0.9214 & 0.9883 & 10.2\% & 10min\\
         SV4D~\cite{sv4d} & 0.9372 & 0.9904 & 8.1\% & 15min \\
         \textbf{Ours} & \textbf{0.9505} & \textbf{0.9912} & \textbf{59.4\%} & 10min \\
         \bottomrule
    \end{tabular}
    \vspace{-5mm}
\end{table}

\topic{Comparison on Video-to-4D.} We also evaluate \themodel on the widely-used Consistent4D~\cite{consistent4d} benchmark to measure the 4D generation quality and temporal consistency. As reported in Tab.~\ref{tab:video-to-4d}, we achieve superior LPIPS~\cite{lpips}, FVD~\cite{fvd}, and very competitive CLIP results, showing the faithful consistency and visual details of our generated 4D contents.
\begin{table}[H]\scriptsize
    \renewcommand{\arraystretch}{1.2}
    \vspace{-3mm}
    \centering
    \caption{Quantitative comparison on Video-to-4D generation.}
    \label{tab:video-to-4d}
    \vspace{-3mm}
    \setlength\tabcolsep{8pt}
    \fontsize{6}{6}\selectfont
    \begin{tabular}{p{2.4cm}<{\centering} | p{1.2cm}<{\centering} p{1.2cm}<{\centering} p{1.2cm}<{\centering}}
        \toprule
        \textbf{Method} & \textbf{LPIPS} $\downarrow$ & \textbf{CLIP} $\uparrow$ & \textbf{FVD} $\downarrow$ \\
         \midrule
         Consistent4D~\cite{consistent4d} & 0.160 & 0.87 & 1133.93 \\
         STAG4D~\cite{stag4d} & 0.126 & 0.91 & 992.21 \\
         DreamGaussian4D~\cite{dreamgaussian4d} & 0.160 & 0.87 & - \\
         4DGen~\cite{4dgen} & 0.160 & 0.87 & - \\
         SV4D~\cite{sv4d} & 0.118 & \textbf{0.92} & 732.40 \\
         \textbf{Ours} & \textbf{0.112} & \textbf{0.92} & \textbf{625.30} \\
         \bottomrule
    \end{tabular}
    \vspace{-5mm}
\end{table}

\topic{Comparison on Text-to-4D.} To evaluate our language-guided motion generation results, we compare with text-to-4D baselines Animate124~\cite{animate124} and 4D-fy~\cite{4dfy} using six Animate124 examples. Specifically, we provide each model with the same 10 prompts generated by GPT4~\cite{gpt4}. It takes 5 hours to generate one Animate124 and 12 hours for one 4D-fy instance, whereas \themodel generates text-guided 3D motions \textit{within seconds} in a single forward pass. We compute the average of the largest 20\% optical flows~\cite{raft} within the instance mask as motion amplitude and conduct a user study to measure the text alignment \& motion diversity. As reported in Tab.~\ref{tab:text-to-4d}, \themodel can efficiently generate more diverse, noticeable, and high-quality text-guided 3D motions.
\begin{table}[H]\scriptsize
    \renewcommand{\arraystretch}{1.2}
    \vspace{-3mm}
    \centering
    \caption{Quantitative comparison on Text-to-4D generation.}
    \label{tab:text-to-4d}
    \vspace{-3mm}
    \setlength\tabcolsep{8pt}
    \fontsize{6}{6}\selectfont
    \begin{tabular}{p{1.4cm}<{\centering} | p{1.6cm}<{\centering} | p{1.5cm}<{\centering} p{1.5cm}<{\centering}}
        \toprule
        \textbf{Method} & \tiny{\textbf{Motion Amplitude}} $\uparrow$ & \tiny{\textbf{Text Alignment}} $\uparrow$ & \tiny{\textbf{Motion Diversity}} $\uparrow$  \\
         \midrule
         Animate124~\cite{animate124} & 0.428 & 6.3\% & 9.3\% \\
         4D-fy~\cite{4dfy} & 0.254 & 12.5\% & 15.6\% \\
         \textbf{Ours} & \textbf{4.319} & \textbf{81.3\%} & \textbf{75.0\%} \\
         \bottomrule
    \end{tabular}
    \vspace{-3mm}
\end{table}

\subsection{Applications}
\vspace{-3mm}
\label{subsec:applications}
\topic{Latent Space Motion Interpolation.} To demonstrate the completeness and continuity of our learned motion embedding, we visualize the decoder's output when interpolating between pairs of motions in the latent space, as shown in \cref{fig:interpolation}. The results indicate that the embedded motion latent codes capture meaningful and common motion patterns of the object, which can be effectively linearly interpolated to generate novel, coherent 3D motions.

\topic{Language-Guided Motion Generation.} Text-to-Motion offers a more user-friendly approach to interactive motion generation. We adopt GPT to generate 300 motion prompts and cluster them into 60 (50 for training and 10 for evaluation) in the BERT embedding space. We project language into the motion latent code and then generate plausible motion sequences in a feed-forward manner, as shown in \cref{fig:language}.

\topic{Test Motions Reconstruction.} For encoding unseen motions, i.e., those in the held-out test set, \themodel demonstrates strong performance in reconstruction quality and motion alignment, as shown in \cref{fig:test_motion}. Given the multi-view videos from the test set, we optimize the latent code, initialized as a standard Gaussian $\mathcal{N}(0, \boldsymbol{I})$ from scratch, while keeping all other parameters fixed. We only use the reconstruction loss for fine-tuning, achieving convergence within 300 steps.

\subsection{Ablation Study}
\label{subsec:ablation}
We validate the effectiveness of our model’s design choices in Tab.~\ref{tab:ablation}. We observe that both motion representation and motion-oriented optimization are critical. Neural key point-based \textit{motion factorization} contributes to effective motion distribution learning and improves the expressiveness of our system. Furthermore, our \themodel leverages the \textit{latent space} to distinguish diverse motion patterns and model the underlying motion distributions. We also verify the effectiveness of \textit{motion pre-training} in achieving robust and precise motion reconstruction. \textit{Multi-motion joint training} within a single generative model forces the network to capture a \textit{shared} 4D geometric structure and learn a \textit{smooth, continuous} motion space. Thus, the generated 3D motions and 4D contents are robust and less sensitive to high-frequency errors or appearance inconsistency of video supervision, compared with single-motion overfitting. More details and qualitative evidence can be found in the supplementary material.
\begin{table}[H]
    \centering
    \vspace{-3mm}
    \caption{Ablation results on different components of DIMO.}
    \label{tab:ablation}
    \vspace{-3mm}
    \setlength\tabcolsep{8pt}
    \fontsize{6}{6}\selectfont
    \begin{tabular}{p{3.0cm}<{\centering}|p{1.0cm}<{\centering} p{1.0cm}<{\centering} p{1.0cm}<{\centering}}
        \toprule
        \textbf{Method} & \textbf{LPIPS} $\downarrow$ & \textbf{CLIP} $\uparrow$ & \textbf{FVD} $\downarrow$ \\
         \midrule
         \textbf{Full Model} & \textbf{0.126} & \textbf{0.93} & \textbf{587.09} \\
         w/o motion factorization & 0.134 & 0.91 & 851.83 \\
         w/o latent space & 0.163 & 0.87 & 1077.19 \\
         w/o motion pre-training & 0.149 & 0.90 & 890.26 \\
         w/o multi-motion joint training & 0.131 & 0.92 & 693.49 \\
         \bottomrule
    \end{tabular}
    \vspace{-3mm}
\end{table}

\section{Limitations \& Conclusion}
\vspace{-1em}
\label{sec:conclusion}
\topic{Limitations and Future Direction.} \themodel relies on video models for object motion prior distillation, indicating that improvements of these models are critical for enhancing our performance. Also, we currently learn language-guided motion generation in two stages by optimizing latent codes and the language projector separately. Jointly learning these two objectives within a single stage will be a key direction.

\topic{Conclusion.} This paper takes the first step toward diverse 3D motion generation for general objects by learning a motion latent space. We achieve state-of-the-art performance in a wide range of standard settings and benchmarks. We hope this work could help us better understand numerous dynamic objects in our physical world and inspire future research in building a general SMPL-like parametric model.

\vspace{6pt}\noindent{\large\bf Acknowledgement}
\vspace{6pt} \\
\noindent We gratefully acknowledge support by the following grants: NSF FRR 2220868, NSF IIS-RI 2212433, ARO MURI W911NF-20-1-0080, and ONR N00014-22-1-2677.

{
    \small
    \bibliographystyle{ieeenat_fullname}
    \bibliography{main}
}

\end{document}